\documentclass[10pt,twocolumn,letterpaper]{article}

\usepackage{wacv}
\usepackage{times}
\usepackage{epsfig}
\usepackage{graphicx}
\usepackage{amsmath}
\usepackage{amssymb}

\usepackage{booktabs} 
\usepackage{xcolor}
\usepackage{xspace} 
\usepackage{authblk}

\newcommand{\aux}{anchor\xspace}

\newcommand{\main}{main\xspace}
\newcommand{\MTLst}{MTL (+both anchor)\xspace}
\newcommand{\freeze}{\textsc{HeadFreeze}\xspace}

\newcommand{\WIP}[1]{}
\newcommand{\wip}[1]{}

\newcommand{\mc}[1]{\mathcal{#1}}

\newcommand{\mcS}{\mc S}
\newcommand{\mcT}{\mc T}
\newcommand{\y}{\checkmark}

\def\wacvPaperID{630} 

\wacvfinalcopy 

\ifwacvfinal
\def\assignedStartPage{1} 
\fi

\ifwacvfinal
\usepackage[breaklinks=true,bookmarks=false]{hyperref}
\else
\usepackage[pagebackref=true,breaklinks=true,colorlinks,bookmarks=false]{hyperref}
\fi

\ifwacvfinal
\else
\fi

\begin{document}

\title{Task-Assisted Domain Adaptation with Anchor Tasks}

\author[1,2]{Zhizhong Li\thanks{Work done prior to current employment. Partially done during the first author's internship at Snap Inc.}}
\author[3]{Linjie Luo}
\author[2]{Sergey Tulyakov}
\author[2]{Qieyun Dai\protect\footnotemark[1]}
\author[1]{Derek Hoiem}

\affil[1]{Computer Science, University of Illinois, Urbana Champaign, IL, USA \authorcr
  \{\tt zli115, dhoiem\}@illinois.edu}
\affil[2]{Snap Inc., Santa Monica, CA, USA \authorcr
  \tt stulyakov@snap.com, qieyunmarydai@gmail.com}
\affil[3]{Bytedance Inc., Mountain View, CA, USA \authorcr
  \tt linjie.luo@bytedance.com}


\maketitle

\begin{abstract}
Some tasks, such as surface normals or single-view depth estimation, require per-pixel ground truth that is difficult to obtain on real images but easy to obtain on synthetic.  However, models learned on synthetic images often do not generalize well to real images due to the domain shift. Our key idea to improve domain adaptation is to introduce a separate anchor task (such as facial landmarks) whose annotations can be obtained at no cost or are already available on both synthetic and real datasets.  To further leverage the implicit relationship between the anchor and main tasks, we apply our \freeze technique that learns the cross-task guidance on the source domain with the final network layers, and use it on the target domain. We evaluate our methods on surface normal estimation on two pairs of datasets (indoor scenes and faces) with two kinds of anchor tasks (semantic segmentation and facial landmarks).  We show that blindly applying domain adaptation or training the auxiliary task on only one domain may hurt performance, while using anchor tasks on both domains is better behaved. Our \freeze technique outperforms competing approaches, reaching performance in facial images on par with a recently popular surface normal estimation method using shape from shading domain knowledge. 
\end{abstract}

\section{Introduction}

\begin{figure}[t]
  \centering
    \includegraphics[width=0.38\textwidth]{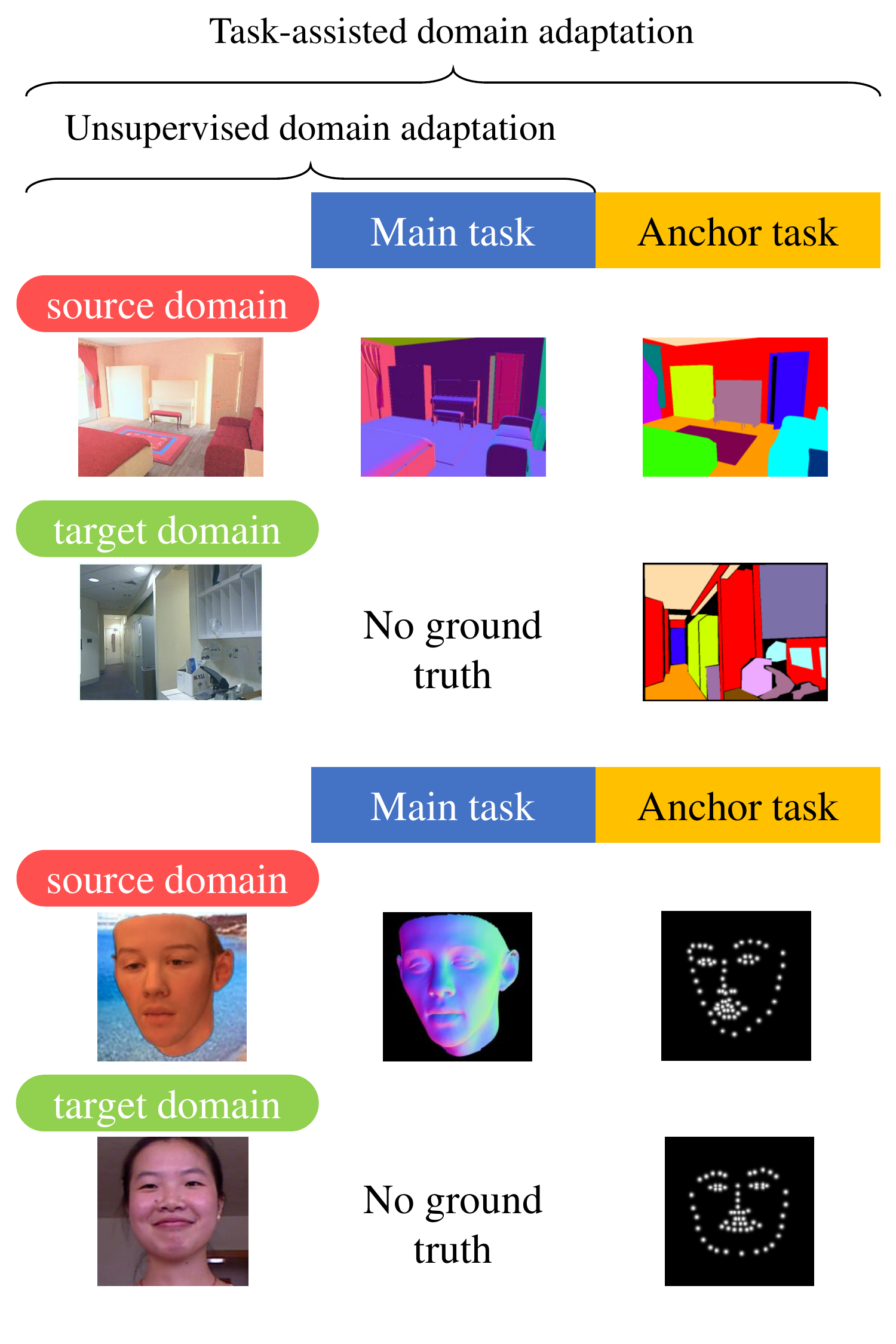}
  \caption{Illustration of our formulation compared to unsupervised domain adaptation. Although target domain \main task labels are hard or expensive to obtain, we can use free or already available labels from an ``\aux'' task to help align the domains with clear correspondence between images and the \aux task label space.\vspace{-2.5mm} }
  \label{fig:intro}
\end{figure}

Collecting annotations is difficult for geometric tasks, such as predicting depth~\cite{song2016suncg,Zhang2017PhysicallyBasedRF}, surface normals~\cite{sengupta2018sfsnet}, and 3D pose~\cite{varol17_surreal}, because it usually requires a specialized device and access to the scene. Synthetic images and their geometric labels are easily generated, but synthetically trained models often do not generalize well to real data.  Unsupervised domain adaptation methods~\cite{tremblay2018training,saito2018maximum,sankaranarayanan2018learning,Hoffman_cycada2017} can help, but they often blindly minimize domain distribution difference~\cite{ganin2015unsupervised,Tsai2019DomainAF,Shrivastava2017LearningFS,Tsai2018LearningTA} even when the ground truth distributions in source and target differ.
How can we better adapt from synthetic to real data?

In this paper, we propose to use {\em anchor tasks} as a guide for improving pixel-level domain transfer of the \emph{main task}. The anchor task is a task labeled on both domains, whose annotations are already available or automatically generated.  For example, in one experiment we improve transfer for surface normal prediction on faces by using facial keypoint detection as an anchor task, and the anchor task ground truth is estimated using an off-the-shelf model.  We propose our \freeze method that first trains the main task and anchor task on synthetic data, then freezes the top few layers and retrain the feature layers to perform the main task on synthetic data and the anchor task on both domains. 
The fully-supervised anchor task provides additional semantic and spatial context information to learn better feature representations for the real images (hence the name ``anchor''), while the frozen predictors leverage learned ``{cross-task guidance}'' so that the main task on the target domain can be guided by the anchor task.

Our idea can be seen as a generalization of existing works that use a closely related auxiliary task to help adaptation. We call this family of methods ``Task-assisted Domain Adaptation'' (TADA).
Prior work~\cite{Gebru2017FineGrainedRI,Yang20183DHP,Inoue2018CrossDomainWO,Fang2018MultiTaskDA} in the TADA family all rely on problem-specific, explicitly defined mappings between the auxiliary tasks and the main tasks (see Section~\ref{sec:related} for details), and thus are restricted to their own task pair and problem settings.
Unlike prior work, we require only that the anchor task has pixel labels, and \freeze is applicable even when the main-anchor relationship lacks an explicit formulation (e.g. between facial keypoints and surface normal map). We demonstrate this with different anchor tasks for the same main task without changing the framework. 
Our experiments focus on geometric tasks with synthetic and real images as the source and target domains, but our approach also applies to other pixel labeling domain transfer problems. 


The anchor task helps domain adaptation in two ways. First, learning shared features for the anchor task on both domains and the main task on the source domain encourages that the features are effective for the main task in the target domain. Second, there may be a multitask learning benefit, if the anchor task and main task have related labels (e.g. ``ceiling'' is always horizontal) or the same features are useful for both tasks.

Our \freeze method strengthens the first benefit, while being simple enough not to require domain knowledge on the task pair.
Specifically, when our network finishes training on the source domain, its final layers have learned not to output unlikely main-anchor prediction pairs (e.g. flat nose, misaligned object edges) but to output likely pairs. We term this knowledge ``\textbf{cross-task guidance}''. 
But this guidance may be ignored if the network overfits to target domain anchor task and outputs unlikely pairs at will. Freezing the final layers fixes the guidance, and ensures the main and anchor task classifiers continue to rely on the same features and the same mapping from feature space to label space.

We evaluate our methods on two pairs of synthetic and real datasets, performing surface normal estimation in indoor scenes and faces, using semantic segmentation and facial landmark as \aux tasks separately. We show the importance of having \aux labels in \emph{both} domains instead of just one, and that \freeze outperforms compared approaches, reaching results in facial images on par with a popular recent model SfSNet~\cite{sengupta2018sfsnet} that leverages a domain-specific illumination model. We also find that, surprisingly, distribution matching adaptation methods can sometimes hurt performance when the label distributions are different between domains, where \freeze's performance is better behaved in our experiments.

In summary, our main contributions are:
\begin{itemize}
\itemsep0em
\item We propose a novel domain adaptation formulation for pixel-labeling main tasks from synthetic to real, using free or readily available anchor task labels. Our formulation is more widely applicable than existing domain adaptation work that leverages auxiliary tasks (but more restricted than unsupervised domain adaptation).
\item We introduce a \freeze technique to further utilize the spatial and contextual cross-task guidance, that can be applied to different pairings of anchor tasks with the same main task.
\end{itemize}

\section{Related work}
\label{sec:related}

\textbf{Domain adaptation methods using auxiliary tasks that are constrained to specific task pairs} (TADA methods). An emerging line of work recently is using multi-task learning or weakly supervised learning to help unsupervised domain adaptation. Gebru \etal~\cite{Gebru2017FineGrainedRI} adapt fine-grain classification between an easy domain and in-the-wild images with the help of classes' attributes. They adapt a consistency loss between attributes and classes and domain adaptation losses from Tzeng \etal~\cite{Tzeng2015SimultaneousDT}. Yang \etal~\cite{Yang20183DHP} adapts lab-environment 3D human pose estimation for in-the-wild data with only 2D pose ground truth, by jointly training on 2D and 3D labels and aligning domains with a GAN-based discriminator. Fang \etal~\cite{Fang2018MultiTaskDA} adapts a robot grasping application from simulation to real images, and from the indiscriminate grasping task to instance-specific grasping. They perform joint training on all existing labels and the optional input of the instance mask. Inoue \etal~\cite{Inoue2018CrossDomainWO} adapts image object detection to paintings by generating pseudo-labels which are filtered using auxiliary image-level labels.

Our paper has two major differences from these prior work. (1) All prior work have very application-specific constraint formulation on the their task-pair relationship, making them inapplicable to nearly all other tasks. Gebru \etal~\cite{Gebru2017FineGrainedRI} assumes the auxiliary and main annotation to have a known linear relationship. Yang \etal~\cite{Yang20183DHP} constraints the 2D pose and 3D pose to use the same 2D pose output layer. Fang \etal~\cite{Fang2018MultiTaskDA} assumes both tasks' output are both binary prediction, share the same output neuron, and the tasks are differentiated by an extra input. Inoue \etal~\cite{Inoue2018CrossDomainWO} must use a hard-coded procedure to filter erroneous outputs of the \main detection task using the \aux task classification labels. In contrast, our work requires only that the two tasks' annotations are spatial, without any constraint on the output layers or the loss of each task -- a much weaker assumption -- and models the cross-task guidance without hard-coded domain knowledge. Our experiments show that the TADA formulation helps task transfer scenarios beyond these task-specific designs with explicit task relations. (2) We focus on tasks with pixel-wise outputs, such as surface normal estimation or keypoint detection (in the form of heatmaps for each keypoint).

\textbf{Weakly supervised learning}~\cite{Zhou2017ABI,Hu2018LearningTS,Lei2017WeaklySI} uses a weaker label to help infer a stronger label, e.g. when inexact coarse category are provided to help fine-grain classification. These methods are not concerned with domain adaptation, and are similar to our method only in the usage of an auxiliary task with available annotations.

\textbf{Unsupervised Domain Adaptation}~\cite{saito2018maximum,sankaranarayanan2018learning,Hoffman_cycada2017,Shrivastava2017LearningFS,ganin2015unsupervised} is similar to our formulation without the \aux task. Especially worth mentioning is Tsai \etal~\cite{Tsai2018LearningTA}. Instead of matching feature space distributions, they adapt the structured output space to have a similar distribution between domains. This is done by applying a GAN-based domain confusion loss over the output from the two domains, and optionally, the feature space as well. Our method additionally uses the \aux task to help on top of these methods, achieving a more fine-grained adaptation with the correspondence the \aux brings. We experimentally show that UDA may hurt performance due to systematic domain difference, while ours is more robust.

\textbf{Semi-supervised Domain Adaptation}, on the other hand, assumes a small number of target domain samples are labeled, compared to our assumption that a task with free or available labels exists for both domains.  Castrejon~\etal~\cite{castrejon2016learning} have a variant similar to \freeze, but require \main task supervision, contrary to our \aux task idea. This is a separate research direction orthogonal with ours.

\textbf{Combining Multi-task learning and Domain Adaptation}~\cite{Tzeng2015SimultaneousDT,Fourure2017MultitaskML,Xu2018PADNetMG} is topically similar. Besides the TADA works we mentioned earlier, most assume all tasks' labels from the target domain are available, and some still requires specially designed losses or constraints for the task pair (e.g. Tzeng \etal~\cite{Tzeng2015SimultaneousDT} constrains the all tasks to be classification tasks). Whether their formulation are still effective in our unsupervised case is beyond the scope of our paper.

\textbf{Transfer Learning} (e.g. Taskonomy~\cite{Zamir2018TaskonomyDT}), including Multi-task Learning~\cite{Caruana1997} (e.g. UberNet~\cite{Kokkinos2017UberNetTA}) and Meta-transfer Learning (e.g. MAML~\cite{Finn2017ModelAgnosticMF}), are methods that use knowledge learned from one task to help another. Most methods assume all tasks are in the same domain or ignore the domain difference (e.g. Taskonomy~\cite{Zamir2018TaskonomyDT}, UberNet~\cite{Kokkinos2017UberNetTA}), and some assume one task per domain or dataset (e.g. Liu \etal~\cite{Liu2015RepresentationLU}). Our idea makes use of the knowledge in one task to help another, but we are more interested in how having the \aux task knowledge in \emph{both} domains can help domain adaptation instead. Compared to prior methods, we empirically show that the \aux task is needed in \emph{both} domains to bridge the domain gap. 

Among these, UberNet~\cite{Kokkinos2017UberNetTA} has similar formulation with our \MTLst ablation, but without using an \aux task shared by all samples. The paper also ignores any domain difference, and only focuses on tasks in datasets where it has supervision, making it irrelevant to domain adaptation.

Some other methods consider performing the same task on different domains as multitask learning~\cite{Long2017LearningMT}, but in our formulation of multitask learning (performing tasks that have conceptually different labels) they are performing semi-supervised domain adaptation instead.

\textbf{Modeling output spatial structure}~\cite{Mostajabi2018RegularizingDN,Tsai2018LearningTA,Tsai2019DomainAF} is related to how we preserve the cross-task guidance between two tasks' outputs. Mostajabi \etal~\cite{Mostajabi2018RegularizingDN} regularizes semantic segmentation by training an autoencoder on the \emph{semantic labels}, and force the network to use the fixed decoder to output its prediction. We are inspired by these ideas, but are focused on how \emph{two} tasks' output spaces interact, and generalizing across domains.

\section{Method}
\label{sec:method}

\begin{figure}[t]
  \centering
    \includegraphics[width=0.42\textwidth]{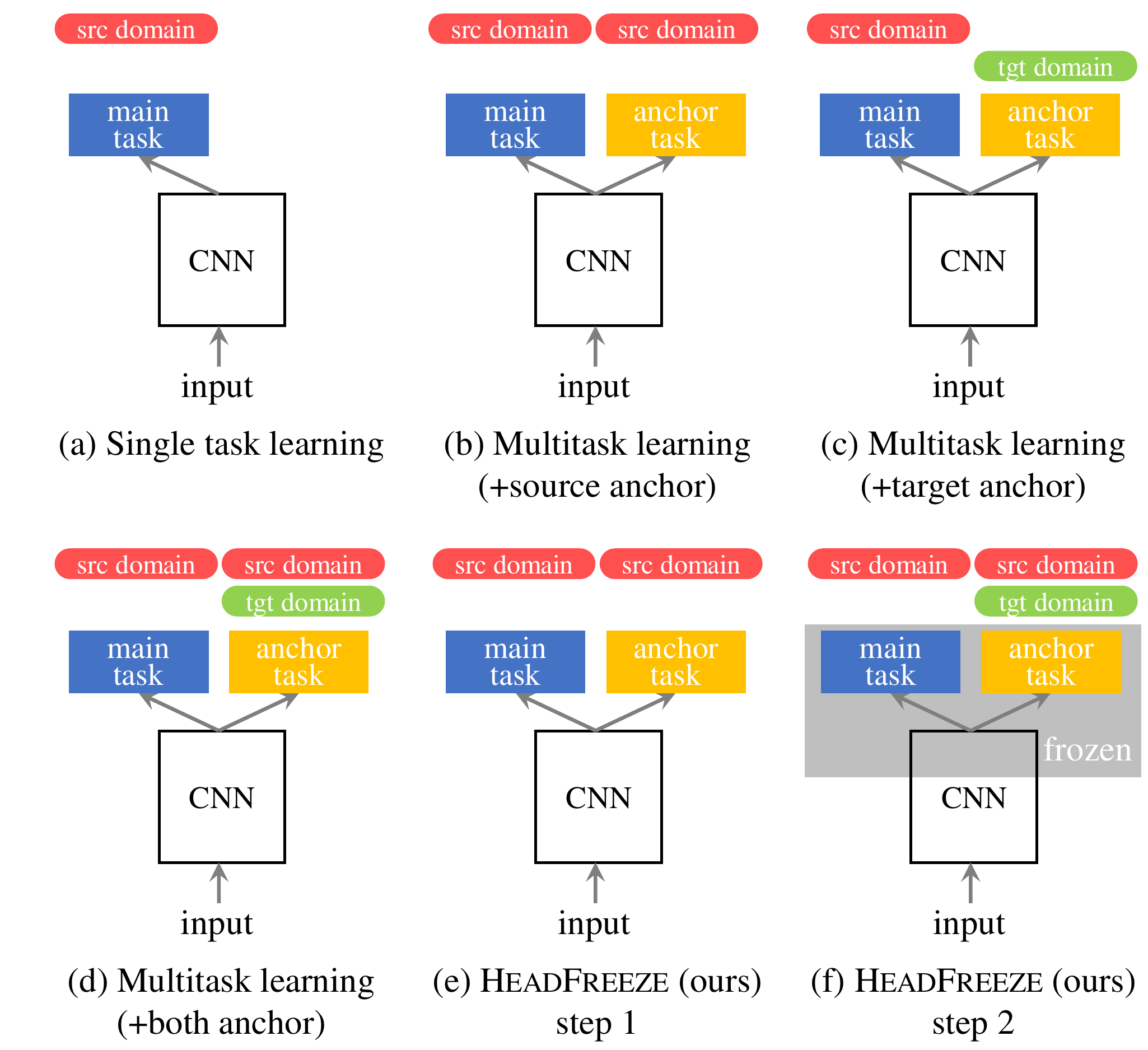}
  \caption{Illustration of various compared methods and their training label usage. TADA methods (d-f) uses the \aux task on \emph{both} domains to establish clear correspondence in the \aux task annotation space. Our \freeze method first trains only on the source domain, and then freezes the final network layers to consolidate the learned cross-task spatial and contextual guidance in the output.}
  \label{fig:method}
\end{figure}

To formulate our Task-Assisted Domain Adaptation (TADA), we first start from a brief review of Unsupervised Domain Adaptation (UDA). In UDA, we have labeled data in the source domain $(x_\mcS, y_\mcS) \in \mcS$, and unlabeled data in the target domain $(x_\mcT, y_\mcT) \in \mcT$. However, only the test set in $\mcT$ may contain labels $y_\mcT$ for evaluation purposes, and in the train set, $(x_\mcT, \varnothing) \in S_{tr}$ is provided. A model, usually with the form of $\hat y = g(f(x))$, is trained on all available data, where $f(\cdot)$ is the network backbone for input-feature mapping, and $g(\cdot)$ is the head for feature-prediction mapping. In this paper, unless otherwise specified, we refer to the networks' second-to-last layer output as the ``\emph{features}''. 

Usually, to reduce the domain gap, features in both domains $f(x_\mcS)$ and $f(x_\mcT)$ are encouraged to follow the same distribution~\cite{ganin2015unsupervised} (although this can also be done in output space $g(f(x_\cdot))$ as well~\cite{Tsai2018LearningTA}). However, it is usually not guaranteed that ground truths $y_\mcS,~y_\mcT$ follow the same distribution. When the ground truths distribute differently, the ideal features and outputs have to distribute differently too. Forcing either of them to distribute similarly would deviate the prediction from the ground truth.

In our Task-Assisted Domain Adaptation scenario, in addition to the \main task, an \aux task is defined for both domains. The domains become $(x_\mcS, y_{\mcS m}, y_{\mcS a}) \in \mcS$, and $(x_\mcT, y_{\mcT m}, y_{\mcT a}) \in \mcT$. Here, $m$ and $a$ stand for the \main and \aux tasks. In the train set of $\mcT$, only $(x_\mcT, \varnothing, y_{\mcT a}) \in T_{tr}$ is provided, while $y_{\mcT m}$ is unknown or unavailable. A model, usually with the form of $\hat y_m = g_m(f(x)),~\hat y_a = g_a(f(x))$ is trained on those data, where $g_m(\cdot)$ and $g_a(\cdot)$ are sub-modules specific to each task. In this work, we focus on the popular formulation above where the two tasks share the same network backbone $f(\cdot)$.

In this work, we consider that the \aux task exists solely to aid the learning of the \main task. We evaluate only on the target domain \main task, not on the \aux. If the \aux task is important, one can always train a separate model for it using a variety of transfer learning methods.

\subsection{MTL + anchor for effective feature learning}
When prior work has performed Multi-task Learning (MTL), either all tasks are assumed to be in one domain, or one task is available in each domain (\main in $\mcS$, \aux in $\mcT$). Formally, there can be three supervised losses in the TADA scenario:
\begin{align}
    \mc L_{\mcS m} &= \mc L_m \left( y_{\mcS m}, \hat y_{\mcS m} \right), \\
    \mc L_{\mcS a} &= \mc L_a \left( y_{\mcS a}, \hat y_{\mcS a} \right), \\
    \mc L_{\mcT a} &= \mc L_a \left( y_{\mcT a}, \hat y_{\mcT a} \right).
\end{align}
In prior work, a multitask learning loss may only comprise of two of the three:
\begin{align}
    \mc L_\text{MTL (+src anchor)} &= \mc L_{\mcS m} + \lambda\mc L_{\mcS a}, \text{ or}\label{eq:mtlsrc}\\
    \mc L_\text{MTL (+tgt anchor)} &= \mc L_{\mcS m} + \lambda\mc L_{\mcT a},\label{eq:mtlsmta}
\end{align}
for ``everything in source'' and ``one task per domain'' respectively. We instead use the alternative baseline -- \MTLst, which simply uses all the supervised losses. 
\begin{equation}
    \mc L_\text{\MTLst} = \mc L_{\mcS m} + \lambda\mc L_{\mcS a} + \lambda\mc L_{\mcT a}.\label{eq:mtla}
\end{equation}
Differences between these formulations are illustrated in Figure~\ref{fig:method}. 

We suggest two ways of choosing the anchor task and obtaining its annotations. (1) Some anchor annotations can be freely obtained, e.g. from very robust estimators that work across most domains, such as facial keypoint detectors. (2) Some anchor tasks can be popular tasks and already have labels in many datasets, such as semantic segmentation. It should be chosen so obtaining it is much easier than the main task annotation.

It may be tempting to hypothesize that the baseline losses in Eq.~\ref{eq:mtlsrc},~\ref{eq:mtlsmta} will be enough for the TADA scenario, and that collecting \aux labels on both domains is not necessary. Maybe in Eq.~\ref{eq:mtlsrc} the multitask learning aspect can already improve model generalization, and in Eq.~\ref{eq:mtlsmta} the network is trained on the target domain, so it may be forced to adapt to perform well on the \aux task. One can also add an unsupervised domain adaptation loss to reduce the domain gap. We experimentally show that these baselines underperform \MTLst and degrade performance.

In addition to any of these supervised losses, an unsupervised domain adaptation loss can be added. For example, adversarial losses (a.k.a. GAN losses) on the features or the output space are used in prior work~\cite{ganin2015unsupervised,Tsai2018LearningTA} with a discriminator network $d(\cdot)$ trained in a mini-max fashion:
\begin{equation}
    \min_{f,g}\max_{d} \mc L(f,g) + \lambda_\text{adv}\mc L_\text{adv}(f,g,d).
\end{equation}
We refer our readers to the prior work~\cite{ganin2015unsupervised,Tsai2018LearningTA} for the exact formulation of $\mc L_\text{adv}$.

\subsection{\freeze for preserving cross-task guidance}

Building on \MTLst, we further propose our \freeze method to leverage the cross-task guidance that can be used to guide the target domain \main task based on the target \aux task.

The final layers of a trained multitask network can can be seen as a decoder from its input feature space to the joint label space of the two tasks. 
When we train these layers on the source domain to convergence, they have only learned to predict output pairs for main and anchor tasks that are \emph{contextually and spatially coherent}, and have never learned to output incoherent pairs (such as misaligned object edges or shapes between tasks, and contradictory outputs like vertical ceilings or flat noses). We assume that the final layers can incorporate this coherency knowledge, and are more likely to predict coherent outputs. It follows that the coherency knowledge can act as a cross-task guidance, so training on target anchor task improves the target main task by ruling out incoherent predictions.

However, it is possible that the model overfits to the target domain anchor task, ignores or forgets any cross-task guidance learned in the source domain. We force the cross-task guidance to persist across domains with \freeze.
We first train the multitask network on the source domain, using Eq.~\ref{eq:mtlsrc}. When it approaches convergence (or just before it overfits to one of the tasks), we freeze the parameters of its final layers. We then train only the lower layers jointly on all available labels using Eq.~\ref{eq:mtla}, forcing their output to go through the pre-trained final layers. See Figure~\ref{fig:method}~(e,f) for an illustration. This procedure can be trained end-to-end by modifying the loss and optimizer's list of variables after the convergence of the first step, which is easy to do in modern frameworks such as PyTorch~\cite{paszke2017automatic}.

For implementation details such as network structure and the number of layers frozen, please see Section~\ref{sec:implement}.

\section{Experiment setup}
We validate our methods and claims on two sets of experiments, facial images and indoor scenes, both adapting from synthetic data to real images -- our motivating scenario. 

\subsection{Facial images}
We perform facial surface normal estimation as the \main task, and for the \aux task we choose 3D facial keypoint detection with automatically generated ground truth. Intuitively, 3D keypoints can inform surface information, and thus is a good form of guidance. As 3D keypoints can currently be reliably generated by methods that generalize well across domains, we use this to show whether \emph{free} \aux task labels can be helpful for another label-deprived task.

We adapt from synthetic data generated by Sengupta \etal~\cite{sengupta2018sfsnet} (``SfSsyn''), using 3DMM models~\cite{Blanz1999AMM}. The dataset provides facial images with surface normal ground truth, with synthetic faces both frontal and looking to the side. We change the reference frame of the surface normal to camera coordinates to follow the definition of all other datasets. 

For the target domain, we use real data from FaceWarehouse~\cite{Cao2014FaceWarehouseA3} (``FaceWH''). The dataset provides facial models fitted using a morphable model followed by a laplacian-based mesh deformation without any PCA reduction, so the surface normals rendered from them are both clean and faithful to the raw RGBD scan.

None of these two datasets provide an official split. We split the subjects (separated by dataset folders) into 70\% for training, and 15\% each for validation and test.

We obtain the anchor annotations for free. On both datasets, we use state-of-the-art Bulat \etal~\cite{bulat2017far} to extract both 3D keypoints and 2D keypoints using their separate models. 3D keypoints are used as \aux training ground truth. We compute the facial region mask from the 2D keypoints for performing evaluation, which is a standard practice in facial surface normal estimation~\cite{Trigeorgis2017FaceN,sengupta2018sfsnet}.

During training, we use the standard losses for both tasks: for surface normal estimation, cosine loss (see~\cite{Trigeorgis2017FaceN}); for 3D keypoint detection, a heatmap regression for the 2D positions, and a vector regression for depth (see~\cite{bulat2017far}). During evaluation of surface normal, we use five metrics in the literature. Specifically, the angular difference between predicted 3D surface normal and the ground truth is treated as the error and computed for each pixel. Then we aggregate the root mean square angular error (RMSE), mean of the error (Mean), median of the error (Median), and percentages of pixels with errors below 11.25$^\circ$ and 30$^\circ$. Only valid regions are considered, so we ignore pixels outside the face or where there is no ground truth (e.g. where depth is missing and surface normal cannot be correctly estimated). 

\subsection{Indoor scenes}
We again perform surface normal estimation as the \main task, but use semantic segmentation as the \aux task to demonstrate, since semantic segmentation has annotations available across many datasets. The semantic boundaries can inform discontinuities in surface normal space, and some categories such as ceilings have very constrained normal directions. Other categories with no fixed shape or expected direction can be hard to improve.

We adapt from the SUNCG dataset~\cite{Song2017SemanticSC} with physically-based rendering~\cite{Zhang2017PhysicallyBasedRF}, which provides images, semantic segmentation, and surface normal ground truth. We use NYUdv2~\cite{Silberman2012IndoorSA} as the target domain, with additional surface normal estimated from depth by Ladicky \etal~\cite{Ladicky2014DiscriminativelyTD}. We only use the labeled portion of the dataset.

SUNCG is large, so we use a 90\%-5\%-5\% split for train, validation, and test. We use NYUdv2's official split. Normal estimation loss and metrics are the same as before, and semantic segmentation is trained using cross-entropy.

\begin{table*}[t]
\centering
\resizebox{0.98\textwidth}{!}{
\begin{tabular}{@{\extracolsep{4pt}} l c@{} c@{}c@{}c @{} c @{} c @{} c @{} c @{} c @{~~} c @{} c @{} c @{} c @{} c  @{} }
\toprule
  &&&&& \multicolumn{5}{c}{Faces:~~SfSsyn$\rightarrow$FaceWH}
  & \multicolumn{5}{c}{Indoor:~~SUNCG$\rightarrow$NYUdv2}
 \\ \cline{6-10} \cline{11-15} \\[-0.8em]
   &$y_{\mcS m}$&$y_{\mcS a}$&$y_{\mcT m}$&$y_{\mcT a}$& $<11.25^\circ$  & $<30^\circ$  & RMSE  & Mean  & Median & $<11.25^\circ$  & $<30^\circ$  & RMSE  & Mean  & Median \\
\midrule
  Baseline                &\y&  &  &  & 0.424 & 0.929 & 17.8 & 14.8 & 12.8                                              & 0.298 & 0.683 & 33.5 & 25.8 & 18.8 \\
  Baseline+DA             &\y&  &  &  & {0.456} & 0.937 & 17.2 & 14.2 & {12.1}                            & \textbf{0.316} & 0.703 & 33.3 & 25.2 & \textbf{17.6} \\ \midrule
  \freeze (ours)     &\y&\y&  &\y& \textbf{0.519} & \textbf{0.954} & \textbf{15.8} & \textbf{12.9} & \textbf{10.9} & {0.301} & {0.708} & \textbf{31.8} & {24.6} & {18.0} \\
  \freeze (ours)+DA&\y&\y&  &\y& {0.455} & 0.935 & 17.2 & 14.2 & {12.1}                            & \textbf{0.316} & \textbf{0.715} & 32.0 & \textbf{24.4} & \textbf{17.4} \\
\midrule
  Oracle             &\y&\y&\y&\y& {0.907} & {0.995} & {7.8} & {6.2} & {5.2}                                       & {0.340} & {0.734} & {30.4} & {23.1} & {16.5} \\
  SfSNet~\cite{sengupta2018sfsnet} &--&--&--&--& 0.495* & 0.965 & 15.2 & 12.9* & 11.3*                             \\
\bottomrule
\end{tabular}
}

  \caption{Comparison in our two experimental settings. Unsupervised domain adaptation with Tsai \etal~\cite{Tsai2018LearningTA} is shown for indoor scenes, and with Ganin \etal~\cite{ganin2015unsupervised} shown for faces, whereas the other combinations underperform (see supplemental). Our \freeze method is comparable to surface normal estimated from SfSNet, without the use of a lighting model. \freeze+DA performs closest to oracle in indoor scene, but domain adaptation methods fail to improve \freeze for faces. Statistical significance computed from 3 runs. (*) denotes a method with domain knowledge performs \emph{equal to or worse} than our best performing method.}
  \label{tab:main}
\end{table*}

\begin{table*}[t]
\centering

\resizebox{0.98\textwidth}{!}{
\begin{tabular}{@{\extracolsep{4pt}} l c@{} c@{}c@{}c @{} c @{} c @{} c @{} c @{} c @{~~} c @{} c @{} c @{} c @{} c  @{} }
\toprule
  &&&&& \multicolumn{5}{c}{Faces:~~SfSsyn$\rightarrow$FaceWH}
  & \multicolumn{5}{c}{Indoor:~~SUNCG$\rightarrow$NYUdv2}
 \\ \cline{6-10} \cline{11-15} \\[-0.8em]
   &$y_{\mcS m}$&$y_{\mcS a}$&$y_{\mcT m}$&$y_{\mcT a}$& $<11.25^\circ$  & $<30^\circ$  & RMSE  & Mean  & Median & $<11.25^\circ$  & $<30^\circ$  & RMSE  & Mean  & Median \\
\midrule
  Baseline                &\y&  &  &  & 0.424 & 0.929 & 17.8 & 14.8 & 12.8                                              & 0.298 & 0.683 & 33.5 & 25.8 & 18.8 \\
  MTL (+src anchor)            &\y&\y&  &  & 0.409 & 0.935 & 17.7 & 14.9 & 13.1                                              & 0.280 & 0.666 & 34.1 & 26.6 & 19.8 \\
  MTL (+tgt anchor)           &\y&  &  &\y& 0.162 & 0.791 & 24.3 & 21.8 & 20.4                                              & 0.260 & 0.662 & 32.9 & 26.2 & 20.6 \\
  \MTLst             &\y&\y&  &\y& 0.492 & \textbf{0.953} & 16.0 & 13.3 & 11.4                                     & 0.275 & 0.675 & 32.4 & 25.7 & 19.8 \\
  \freeze (ours)     &\y&\y&  &\y& \textbf{0.519} & \textbf{0.954} & \textbf{15.8} & \textbf{12.9} & \textbf{10.9} & \textbf{0.301} & \textbf{0.708} & \textbf{31.8} & \textbf{24.6} & \textbf{18.0} \\
\bottomrule
\end{tabular}
}
\caption{Ablation studies. See Figure~\ref{fig:method} for each method's formulation. Other MTL baselines underperform while \MTLst outperforms, indicating the importance of shared \aux tasks. Our \freeze technique further boosts \MTLst performance. Statistical significance computed from 3 runs.}
  \label{tab:ablation}
\end{table*}

\subsection{Compared methods}
Since we address the domain adaptation problem, we compare to unsupervised adaptation methods that applies either a multi-level version of Ganin \etal~\cite{ganin2015unsupervised} or state-of-the-art Tsai \etal~\cite{Tsai2018LearningTA}, either on the single task model (DA), or on our method for further improvement (\freeze+DA). Adversarial training is brittle and not all configurations work too well. We implement our own version and perform hyperparameter tuning, and omit some of the underperforming combinations. 
We also compare to an oracle method that uses both tasks labels on both domains, including the target domain \main task. This gauges how far each method is from fully successful adaptation. 

For facial surface normal, we compare to a recent and popular intrinsic decomposition method SfSNet~\cite{sengupta2018sfsnet}, which produces surface normal based on extra domain knowledge (lighting model for unsupervised learning). We use their released model trained on synthetic data and on unsupervised CelebA~\cite{liu2015faceattributes}, a much larger dataset. This comparison only serves to prove that our method is effective instead of being a controlled experiment, since neither our network structure or external knowledge is similar.

For ablation studies, we compare to baselines shown in Fig.~\ref{fig:method}: single task baseline, multitask with only one domain (MTL (+src anchor)), multitask with source \main task and target \aux task (MTL (+tgt anchor)) as used in prior work such as Liu \etal~\cite{Liu2015RepresentationLU}, and \MTLst.

\subsection{Implementation details}
\label{sec:implement}
Code will be released. We use a ResNet50~\cite{He2016DeepRL} with FPN~\cite{Lin2017FeaturePN} for our network backbone, with the ResNet pre-trained on ImageNet~\cite{ILSVRC15}. We use the variant with 3 upsampling layers with skip connection, and used a deconvolution layer as the output layer for both tasks, making the output 50\% of the input resolution. For \freeze, we freeze the layers after the second upsampling layer, including any skip connection weights. Some tasks require additional non-spatial outputs. A common practice of 3D keypoint estimation~\cite{bulat2017far} is to output a heatmap for projected 2D positions, and a vector for 3D depth. We add a fully-connected branch of 2 layers with 256 hidden units after the global average pooling over the second upsampling layer's output. Batches of the same size are sampled from each domain for each iteration. We choose $\lambda$ so losses from different domains and tasks have similar magnitudes.
For adversarial training and dataset processing, please refer to our supplemental material.

Hyperparameter tuning is hard in TADA, just like in any unsupervised domain adaptation, due to the lack of target domain \main task ground truth in validation. Although by evaluating against available ground truth we can tune most hyperparameters (e.g. stop criteria, learning rate, layers to freeze), some parameters critical to target \main task (e.g. discriminator network complexity, its learning rate and loss weights) may barely cause any change. We empirically find that the discriminator accuracy being very frequently lower than 55\% and good qualitative results (absence of artifacts) are good indicators of successful adaptation, and tune the parameters accordingly.


\section{Results}
\label{sec:results}

\begin{figure*}[t]
  \centering
    \includegraphics[width=0.75\textwidth]{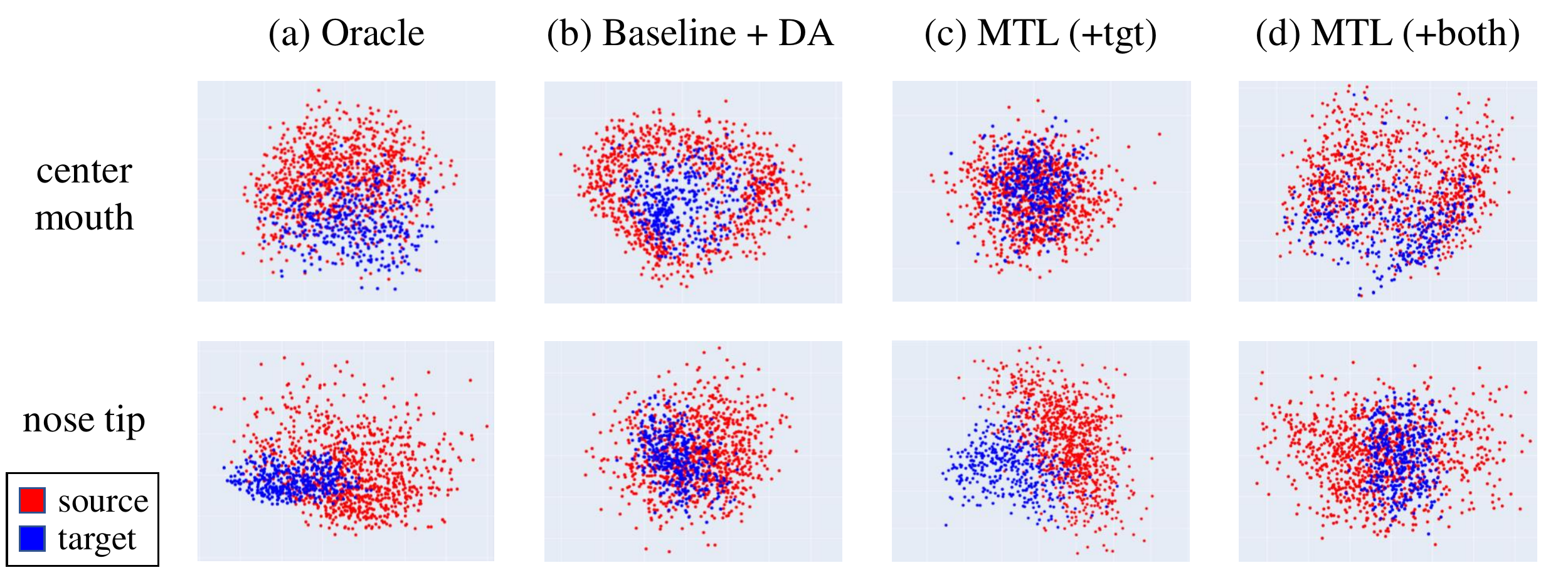}
  \caption{PCA visualization of the subtle differences between methods' feature space at different facial keypoint locations. (a) Oracle does not have fully overlapping domain due to systematic distribution differences. (b) Forcing domains' distributions to be similar can deviate the features from the oracle and hurt performance (top row). (c) Training MTL with one task per domain may encourage using separate feature space regions for different domains (bottom row). (d) \MTLst produces feature distributions slightly more visually similar to the oracle. Disclaimer: the baseline's visualization (not shown) is also similar to the oracle, so this cannot indicate higher performance. Other facial locations may not exhibit observed behaviors as clearly. Best viewed in color.}
  \label{fig:scatter}
\end{figure*}

\begin{figure*}[t]
  \centering
    \includegraphics[width=0.80\textwidth]{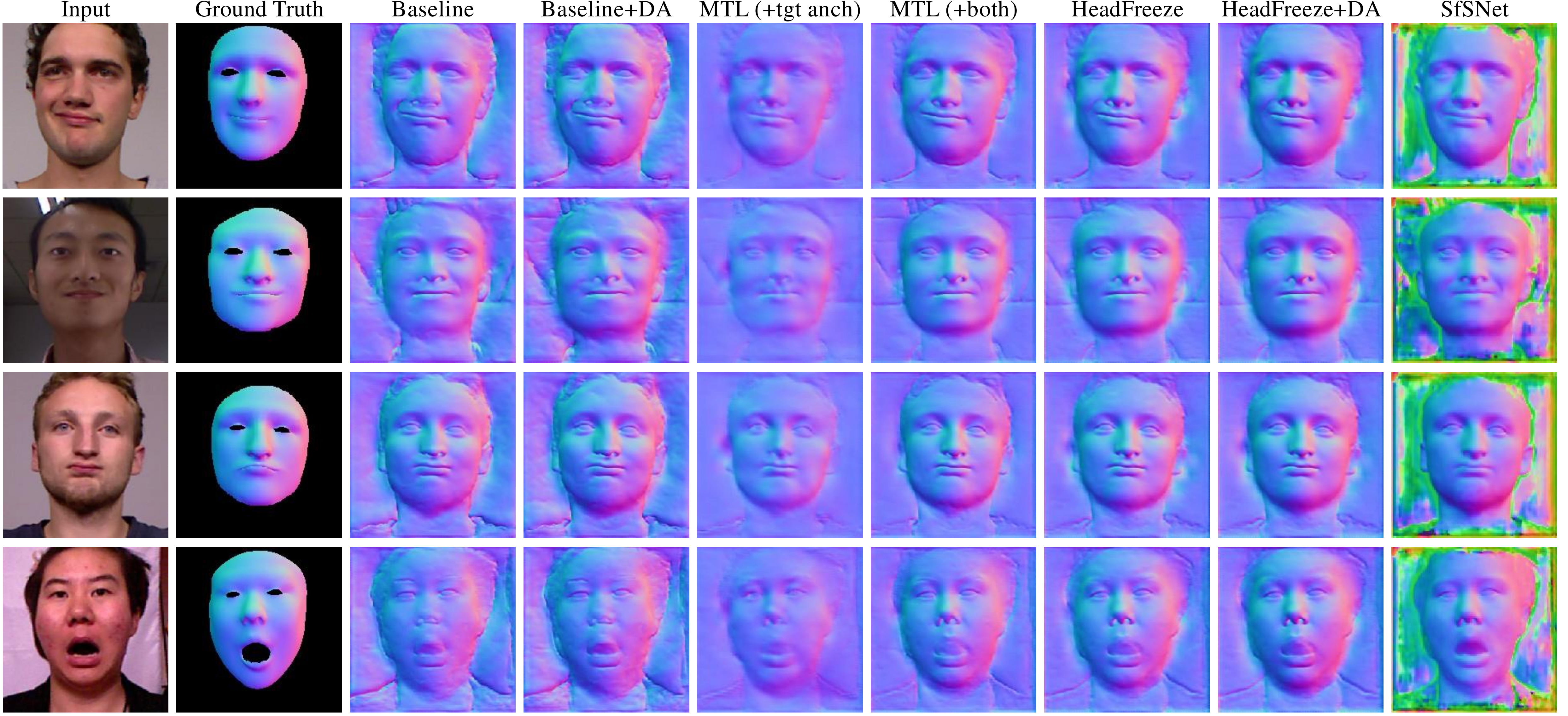}
    \\[1em]
    \includegraphics[width=0.80\textwidth]{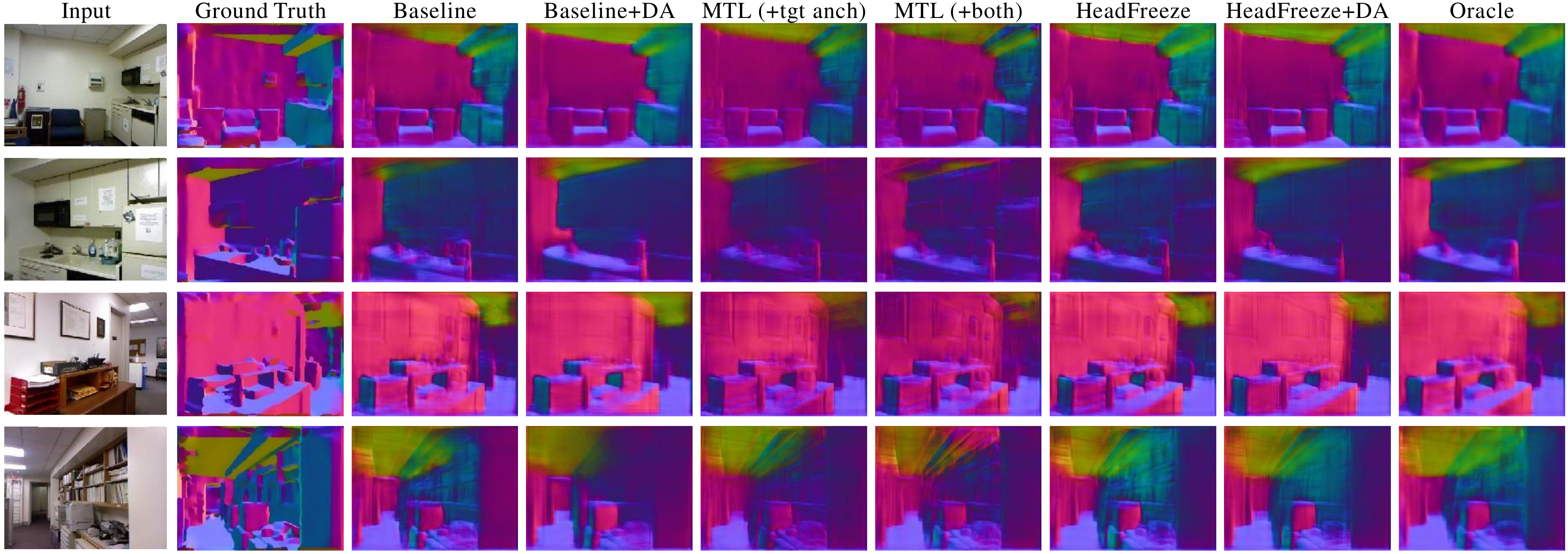}
  \caption{Qualitative results for compared methods. For domain adaptation, Ganin \etal~\cite{ganin2015unsupervised} is shown for facial images (top), and Tsai \etal~\cite{Tsai2018LearningTA} is shown for indoor scenes (bottom). Best viewed in color.}
  \label{fig:qualitative}
\end{figure*}

Table~\ref{tab:main},~\ref{tab:ablation} shows our results and ablation studies.

\textbf{Facial images}. On SfSsyn to FaceWH adaptation, \freeze outperforms the non-adaptation baseline. Adding domain adaptation~\cite{ganin2015unsupervised} does improve baseline results, but still underperforms our \freeze method. \freeze is comparable to the popular SfSNet~\cite{sengupta2018sfsnet}, which underperforms on Median and 11.25$^\circ$ but outperforms on RMSE and 30$^\circ$. However, all methods are still quite far from the oracle method that uses target domain main task annotations.

Perhaps a very surprising observation is that the unsupervised domain adaptation methods added to \freeze would \emph{hurt} performance instead of improving them. In fact, \freeze \emph{without adaptation} is the best method apart from the oracle (and SfSNet). The adaptation puts \freeze at the same level with DA~\cite{ganin2015unsupervised}, eliminating any advantage brought by the \aux task. We have vigorously tuned the adversarial loss hyperparameters, yet still cannot find a configuration that would not hurt performance. In comparison, \freeze (and even \MTLst in Table~\ref{tab:ablation}) work naturally. We analyze the reason for our robustness in Section~\ref{sec:analysis}.

In the ablation study, the baseline and MTL with anchor on either one domain all underperform. Two observations are interesting: (1) MTL (+src anchor) does not perform very differently from the baseline on the target domain, indicating that the effect of multi-task learning is limited here. (2) MTL (+tgt anchor) vastly underperforms the baseline when trained with one task per domain. We hypothesize that despite the network being trained on the target $\mcT$, the task performed on $\mcT$ is too different, which \emph{encourages} the network to learn very different features for the tasks, harming adaptation. 
\MTLst outperforms other MTL methods, indicating the importance of the \aux task being trained on both domains, affirming our hypothesis. \freeze further improves all criteria by a margin, implying that the cross-task guidance learned in the source domain can be helpful for the target domain as well.

\textbf{Indoor scenes.} We still see both \freeze and domain adaptation~\cite{Tsai2018LearningTA} improve over the non-adaptation baseline, but it is inconclusive whether \freeze outperforms domain adaptation~\cite{Tsai2018LearningTA}. But we observe that \freeze+DA further improves the adaptation-only method, closing much of the gap between baseline and the oracle.

In Table~\ref{tab:ablation}'s ablation study, all MTL variations suffer from negative transfer, i.e. \main task performance degrades as the second task is jointly learned. 
We still observe that \MTLst outperforms other MTL variants, indicating that it has an adaptation effect that other variants do not possess, despite the negative transfer. We also observe that \freeze makes a larger improvement on \MTLst than in the facial experiments. 

\subsection{Further analysis}
\label{sec:analysis}
\textbf{Failure of adaptation and face label distribution.} We analyze why the compared domain adaptation methods fail to improve \freeze in the SfSsyn-FaceWH experiment. After trials and errors, we found that the difference of head pose distributions between domains may be a major contributor. 
We generated a second version of the SfSsyn dataset with only frontal faces (``SfSsyn-front''), with rotation distribution closely following the estimated poses from the target dataset.
We evaluate the domain adaptation methods with SfSsyn-front as the source domain in Table~\ref{tab:frontal} instead, with all methods using the same hyperparameter.

\begin{table}[t]
\resizebox{0.48\textwidth}{!}{
\begin{tabular}{@{\extracolsep{0pt}} l @{ } c @{ } c @{ } c @{ } c @{ } c  @{} }
\toprule
   & $<11.25^\circ$  & $<30^\circ$  & RMSE  & Mean  & Median \\
\midrule
  Baseline & 0.418 & 0.913 & 18.6 & 15.3 & 13.0 \\
  Baseline+DA~\cite{ganin2015unsupervised} & 0.495 & 0.944 & 16.6 & 13.5 & 11.3 \\
  \freeze & 0.550 & 0.958 & 15.2 & 12.4 & 10.4 \\
  \freeze+DA~\cite{ganin2015unsupervised} & \textbf{0.573} & \textbf{0.963} & \textbf{14.7} & \textbf{11.9} & \textbf{10.0} \\
\bottomrule
\end{tabular} 
}
\caption{Facial normal estimation, with SfSsyn-frontal as the source domain, which has head pose distribution similar to FaceWH. In this experiment, domain adaptation~\cite{ganin2015unsupervised} always helps performance, indicating that systematic dataset difference is the reason distribution matching adaptation fails, which \freeze is robust to.}
  \label{tab:frontal}
\end{table}

The trends and conclusions are exactly the same, except that unsupervised domain adaptation always helps, making our \freeze + DA the top method. This experiment indicates that the distributional difference is indeed why adaptations~\cite{ganin2015unsupervised,Tsai2018LearningTA} fail. We conclude that while these prior works are effective, they would \emph{hurt} performance when domain ground truths are differently distributed, whereas our methods are more robust to such differences. While these differences may sometimes be easily eliminated in data synthesis procedures, other times they may be expensive to eliminate, or difficult to pinpoint.

\textbf{Impact on feature space.} To better understand the impact of different methods on the feature distribution, we visualize their feature space for source and target domain. Since features at different spatial locations may encode information differently, we extract the feature at separate facial keypoint locations in the facial experiment. For each location (e.g. nose tip), we perform PCA and obtain the top two components, and visualize them in Fig.~\ref{fig:scatter}. Please refer to its caption for observations. This experiment resonates with our hypothesis that training MTL (+tgt anchor) with one task per domain would map source and target to different feature space regions, and that blindly matching feature distribution may be suboptimal.

\textbf{Qualitative results} are shown in Figure~\ref{fig:qualitative}. For faces, the synthetic dataset has less facial expressions than FaceWarehouse, so baselines struggle with e.g. open mouths. Unsupervised adaptation~\cite{ganin2015unsupervised} tends to erroneously force the cheeks and nose normals to the side to force the output look like side-facing faces locally. The ground truth is not extremely faithful to the image due to being fitted on RGBD scans, and both our \freeze method and SfSNet~\cite{sengupta2018sfsnet} capture local details better than the ground truth, although SfSNet performs better with open mouths due to their usage of a lighting model on unlabeled real faces. 

For indoor scenes, \freeze improves the performance for shelves, cabinets, and ceilings more effectively than the facial datasets, possibly due to their semantic labels providing much information for their surface normal.

\section{Summary and future work}

We propose a strategy to extend prior Task-assisted Domain Adaptation methods by eliminating the need for task-specific relationship formulations. We use spatial information of a free or already available shared \aux task to align both features between domains and spatial prediction and context between tasks, and propose \freeze to further leverage the cross-task guidance to improve target domain \main task. We show effectiveness and robustness of using \aux tasks against multitask baselines, and \freeze against conventional adaptation methods.

There are many open questions to answer for the effect of \aux tasks. How do we make sure \main task get information from \aux task output directly? Would a design built on PAD-Net~\cite{Xu2018PADNetMG} work? Can we adapt multiple \main tasks from only one \aux task to leverage all the rich labeling of synthetic data? How cheap can the \aux task be made? Can Taskonomy~\cite{Zamir2018TaskonomyDT} help in choosing which \aux task to use? We leave these questions for future work.

\textbf{Acknowledgments}
This work is supported in part by the Office of Naval Research grant ONR MURI N00014-16-1-2007 and a gift from Snap Inc.

{\small
\bibliographystyle{ieee_fullname}
\bibliography{egbib}
}

\end{document}